**Ali ELouafiq**
**Supervised by: Dr. Violetta Cavalli Sforza**
**CSC3309**

**Machine Cognitive Models**

1. **Introduction:**

Through history, the human being tried to relay its daily tasks to other creatures, which was the main reason behind the rise of civilizations. It started with deploying animals to automate tasks in the field of agriculture(bulls), transportation (e.g. horses and donkeys), and even communication (pigeons). Millenniums after, come the Golden age with "Al-jazari" and other Muslim inventors, which were the pioneers of automation, this has given birth to industrial revolution in Europe, centuries after. At the end of the nineteenth century, a new era was to begin, the computational era, the most advanced technological and scientific development that is driving the mankind and the reason behind all the evolutions of science; such as medicine, communication, education, and physics. At this edge of technology engineers and scientists are trying to model a machine that behaves the same as they do, which pushed us to think about designing and implementing "Things that-Thinks", then artificial intelligence was. In this work we will cover each of the major discoveries and studies in the field of machine cognition, which are the "Elementary Perceiver and Memorizer"(EPAM) and "The General Problem Solver"(GPS). The First one focus mainly on implementing the human-verbal learning behavior, while the second one tries to model an architecture that is able to solve problems generally (e.g. theorem proving, chess playing, and arithmetic). We will cover the major goals and the main ideas of each model, as well as comparing their strengths and weaknesses, and finally giving their fields of applications. And Finally, we will suggest a real life implementation of a cognitive machine.

2. **Elementary Perceiver And Memorizer (EPAM):**
    a. **Goals[2]:**

    Having its roots in the nineteenth century, based on nonsense syllables that were firstly introduced by Ebbinghaus in the 1870s, EPAM was developed by the Edward Feigenbaum and Herbert Simon between 1956-1964[7], which aims to learn nonsense material, and has both a short term and a long term memory. That will make it avoid learning of the stimuli based on their meaning, and will push it to use mnemonic techniques to remember the stimuli.

    This model gives also explanation of some human characteristics of learning, which are mainly oscillation, retroactive inhibition, forgetting, stimulus, and response generalization. These elements will be discussed within this work.

b. **Main Ideas:**

EPAM based on paired-association by learning by criterion. This system learning is based on many trials that enable him to memorize the pair of syllables. In each trial EPAM should memorize non sense syllables represented in lists of associated pairs. In a specific example, a syllable is composed of 3 letters, and begins and ends with a consonant. In a paired associates model the first variable is named stimulus and the second is named response. The trails are repeated many times, through many trials EPAM will learn a lot of material which may push it to remember fast the non confusing syllables but it may force him to forget other material.[3]

   i. **Behavior of paired-based verbal learning:[2]**

Paired based based learning has a set of behaviors that are similar to humans learning behaviour, which are mainly:

Stimulus and Response generalization:
Which is making confusion when similar stimuli or similar responses are presented.

Oscillation:
Associations that are relevant in several trials may some times be forgotten and then be remembered in later trials.

Retroactive inhibition:
a term borrowed from psychology, in theory of learning[1], which is a form of forgetting memorized information, through a number of trials, but the memory may be refreshed later during the coming trials.

   ii. **How does EPAM Works: The Performance System[2]:**

The Performance system is one of the two main building blocks that drives EPAM, this system is responsible to produce responses to stimulus syllables, in other words it the Performance system who fetch the memory for the learned material.

The Process:
The Performance system receives an input, then it search for it through the *discrimination network*, which gives us a *cue* for the response then the Performance system search again based on the cue to get to the response image node.

In order to perform such operations EPAM encodes stimuli into an internal representation called the *Input code.* Based on many features of the syllables (e.g. openness of the letter), the *Input code* is built, and this set of features should abide to two rules: highly redundant features, and relative to the letters or symbols.

The Discrimination Network

The Discrimination Network is actually a binary search tree which internal nodes are tests of the features and the leafs are cues or response images.

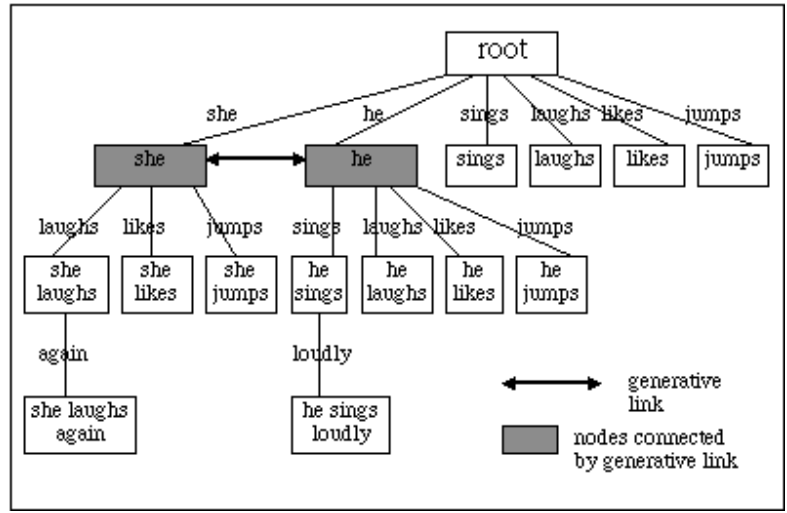

Figure 2.a.ii: Example of discrimination Network [8]

### iii. How does EPAM Works: The Learning System:

The Learning System's main job is to construct the discrimination Network, and it also adds nodes to it when new material is learned.

The cues and images are generated during the learning process, the cue is a hint(leaf) that acts as a new input for the search, this cues are generated based on the responses, and placed in the tree according to the syllable. The response images are constructed based on the response feature.

When a new pair is introduced, the Learning system look for the appropriate place for the syllables, when it finds the place it adds a cue as a leaf. Then the Learning system search for the response a new place to put its image.

When similar cues leads to the same response, the new cue is changed to avoid this confusion.

### c. Strengths and Weaknesses:

One of the major strengths of EPAM is that enables a machine to learn information and memorize it without any reference to the real meaning of the learned material, just based on features of the information. Which enable the machine to solve problems or fetch for solutions in a human behavior.

But the problems in fetching information occur when the set of learned material become large enough, which lead to a rise of the probability of giving wrong answers and overwriting learned material. Also, the classical implementation does not take into consideration *proactive inhibition.* Last but not least, when the information stored become too large, fetching information may lead to long searches in some cases.

### d. Other Applications of Techniques:
EPAM model could be used in several fields of computer science, to solve various world problems, we can denote:

Human Languages: EPAM could be used for syntax correction.

Computer Vision: If we can get from the image set of features that identifies the object that we want to detect , a face for example, then the object recognition will be faster than the classical methods used, such as statistical methods that needs a lot of computation.

It can be used in Thesaurus for finding the synonyms of a certain word, and it will be regardless of the language.

It may also be useful for chat bots, to make the search for the specific answer faster.

## 3. General Problem Solver (GPS):
### a. Goals [3,4]:

In order to get machine solve problems requiring deep levels of intelligence, and trying to develop a theory that explains how human are solving such problems, by Newell, Shaw, and Simon beginning in 1957, developed a new concept and model that tries to reach these goals, which was named the General Problem Solver (GPS). GPS was mainly specialized particularly three task domains crypt-arithmetic, logic; and chess—and they present and evaluate information- processing systems that accurately simulate human thought in these domains.

### b. Main Ideas [3]:
The fact that most of the intellectual activities and problem solving involves two main types of knowledge, in other to come up with a solution:

General Knowledge: knowledge that applies to many problems.
   Example: "If this alternative doesn't work, try the other alternative."
   "If you can't solve the whole problem, divide it and solve partial parts."
Very specific knowledge: that is special to a particular problem.
   Example: "whatever is green is not blue".
   or "(number,1) and (sign,+) and (number,1) => (number,2)".

This main distinction composes the backbone of the GPS, especially that this new system was separating in a clean way a task-independent tier of the system, containing general problem solving mechanisms (General knowledge), from a tier of the system that contains the knowledge acquired of the task environment provided.

i. **Task environment:**

The GPS is supposed to be task-independent, in a way that the problem solving part of the system should not give any information about the kind of task being worked on. Thus, the task dependent knowledge is collected in special data structures named *task environment*.

<u>Tasks:</u> A task is given to the GPS as an *initial <u>object</u>* and as *desired objects*.
<u>Operators:</u> is any activity that changes the state (i.e. *object*) of the system.

ii. **Mean-ends Analysis:**

Mean-ends analysis , one if the major contributions of the GPS, It assumes that the differences between a current object and a desired object can be defined and classified into types and that the operators can be classified according to the kinds of differences they might reduce.

At each stage, GPS selects a single relevant operator to try to apply to the current object. A Depth First Search for a successful sequence is performed as long as the operators chosen are applicable; there the path approaches to the goal. Backtracking (i.e. backup) is used to search other alternatives if the path becomes promising (e.g. a simplification makes a tougher object that are more difficult to simplify than the initial, or previous, object)

iii. **The Goal Structure:**

The goal is a representation of the current list of and the desired list of objects, as well as the history of the trials to change the current object.

Goals are categorized into 3 kinds:
- **Transform** object A → object B.

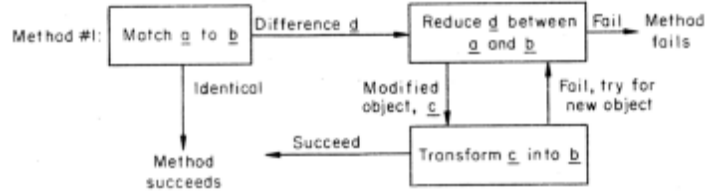

Figure 3.b.iii.1- Flowchart of transform object A into object B
- **Reduce** difference between A(object) and B(object) by modifying object A.

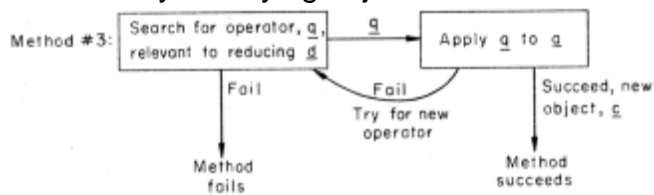

Figure 3.b.iii.2- Reducing the difference
- **Apply** operator **Q** to object **A**.

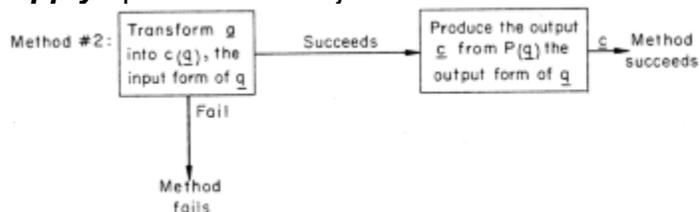

Figure 3.b.iii.3- Flowchart Describing Apply goal

### iv. **The Process [9,10,4,3]:**

The GPS tests various actions (as described above as operators) to look for the one which will take it closer to the goal state. In this way The GPS always chooses the operation that seems to bring it closer to the intended goal. This is similar to the well known tactic in offline searches called hill climbing, since it behaves as if it is always taking a step toward the top of a hill or mountain. As stated in the work Psychology, introduction by Russ Dewey, "Similarly, humans struggle toward long-term objectives "one day at a time."[9] Progress through school can be represented as a problem space. As you complete each course in your curriculum, you take one step closer to the goal of obtaining a degree."

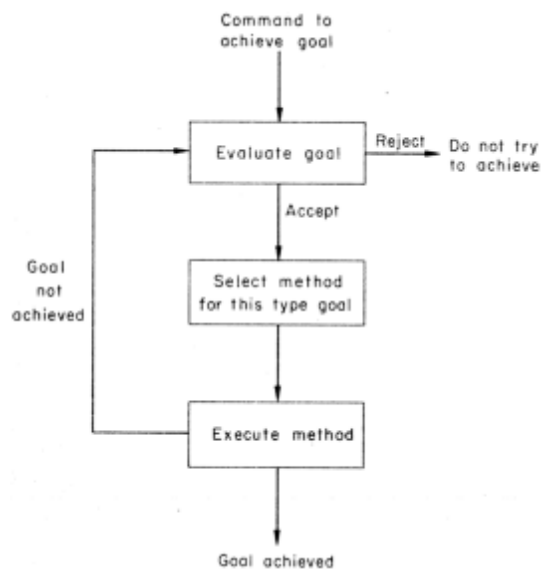

Figure

### c. **Strengths and Weaknesses[10,5,3]:**

Strengths:
- The Ability to perform theorem proving.
- Strong simplification of complex states.
- Simulation of the human process of thinking.
- Giving the path necessary to solve the problem.

Weaknesses:
- Limited memory constraints, since the GPS need large extensions.
- May stack in an infinite loop by looping over complex states (i.e. if he finds a complex state, and tries to find another path that will be also complex)
- Limited types if problems that can be solved by the system, motivated problems needs a large definition of their rules, thus

> we may pay the overhead of performance getting (or not getting at all) a solution after days of computations.

   d. **Other Applications of Techniques:**

   The General problem solver can be used in First Order Logic; it will be easy to define the rules of the system before it begins. Especially that GPS deals with symbolic notations perfectly.
   The General problem Solver can be used as well in trigonometry proving, since this special topic in mathematics has a definite set of rules and limited set of variables (unlike Logic), and it will be used to simplify complex formulas (i.e. trigonometry is one of the most important topics used in optimization problems).
   The GPS can be used in Natural languages processing, by using its reduction behavior to change meaning of the text, by which operators will be grammatical transformations, or thesaurus synonyms searching.

4. **Conclusion**

   EPAM and GPS are considered as one of the most important discoveries in the Artificial Intelligence and in psychology field that tries to simulate human thinking behavior and to solve problems. Each one of them has some limitations, not only the complexity of its execution time and space, but the ability to solve dedicated problems as well, which is the main objective. In machine cognition in general, and in artificial intelligence particularly, having universal systems that solves all kind of problems, or at least covering a large set of them (problems), is remaining a challenge, many trials have been done, but till now our progress is very limited compared to real life intelligence available in nature. Some relate this to the computational limitation of our computers (e.g. the traveling sales man problem requires huge amount of computation, while a normal bee solves it easily in its trip from a flower to another) and others relates it to the algorithms and systems we designed, which are not optimal compared to the natural intelligence. Perhaps never will we reach the singularity.